\documentclass{article} 
\usepackage{conference,times}

\usepackage{amsmath,amsfonts,bm}

\def\eqref#1{equation~\ref{#1}}

\def\1{\bm{1}}

\DeclareMathAlphabet{\mathsfit}{\encodingdefault}{\sfdefault}{m}{sl}
\SetMathAlphabet{\mathsfit}{bold}{\encodingdefault}{\sfdefault}{bx}{n}

\newcommand{\E}{\mathbb{E}}

\DeclareMathOperator*{\argmin}{arg\,min}

\usepackage{hyperref}
\usepackage{url}
\usepackage{amsmath}
\usepackage{amssymb}
\usepackage{amsthm}
\usepackage{enumitem}
\usepackage[capitalize,noabbrev]{cleveref}
\usepackage{url}            
\usepackage{booktabs}       
\usepackage{amsfonts}       
\usepackage{nicefrac}       
\usepackage{microtype}      
\usepackage{xcolor}         
\usepackage{multicol}
\usepackage{multirow}
\usepackage{pifont}
\usepackage{algorithm}
\usepackage{algorithmic}
\usepackage{wrapfig}
\usepackage{comment}
\usepackage{xspace}
\usepackage{arydshln}
\usepackage{subcaption}
\usepackage{caption}
\usepackage{mathtools}
\usepackage{xcolor,colortbl}
\usepackage{arydshln,dashrule}

\usepackage[textsize=tiny]{todonotes}

\newcommand{\x}{\pmb{x}}
\newcommand{\s}{\pmb{s}}
\newcommand{\thet}{\pmb{\theta}}
\renewcommand{\L}{\mathcal{L}}
\newcommand{\T}{\mathcal{T}}
\renewcommand{\S}{\mathcal{S}}
\newcommand{\A}{\mathcal{A}}
\newcommand{\D}{\mathcal{D}}

\newcommand{\alg}{\texttt{PDD}\xspace}

\title{Data Distillation Can Be Like Vodka: Distilling More Times For Better Quality}

\author{Xuxi Chen*\textsuperscript{1},
Yu Yang*\textsuperscript{2},
Zhangyang Wang\textsuperscript{1},
Baharan Mirzasoleiman\textsuperscript{2} \\
{\textsuperscript{1}University of Texas at Austin} 
{\textsuperscript{2}University of California, Los Angeles}
  \\
  \small{\texttt{\{xxchen,atlaswang\}@utexas.edu}}, \small{\texttt{\{yuyang,baharan\}@cs.ucla.edu}}
}

\iclrfinalcopy 
\begin{document}

\maketitle

\begin{abstract}
Dataset distillation aims to minimize the time and memory needed for training deep networks on large datasets, by creating a small set of synthetic images 
that has a similar generalization performance to that of the full dataset. However, current dataset distillation techniques fall short, showing a notable performance gap when compared to training on the original data. 
In this work, we are the first to argue that using just one synthetic subset for distillation will not yield optimal generalization performance. This is because the training dynamics of deep networks drastically change during the training. Hence, multiple synthetic subsets are required to capture the training dynamics at different phases of training. To address this issue, we propose Progressive Dataset Distillation (\alg). \alg synthesizes multiple small sets of synthetic images, each conditioned on the previous sets, and trains the model on the cumulative union of these subsets without requiring additional training time.
Our extensive experiments show that \alg can effectively improve the performance of existing dataset distillation methods by up to $4.3\%$. In addition, our method for the first time enable generating considerably larger synthetic datasets.

\end{abstract}
\renewcommand{\thefootnote}{\fnsymbol{footnote}}
\footnotetext[1]{Equal Contribution.}
\renewcommand{\thefootnote}{\arabic{footnote}}
\vspace{-4mm}
\section{Introduction}

\begin{wrapfigure}{r}{0.45\linewidth}
    \centering
     \vspace{-2em}    
     \includegraphics[width=1\linewidth]{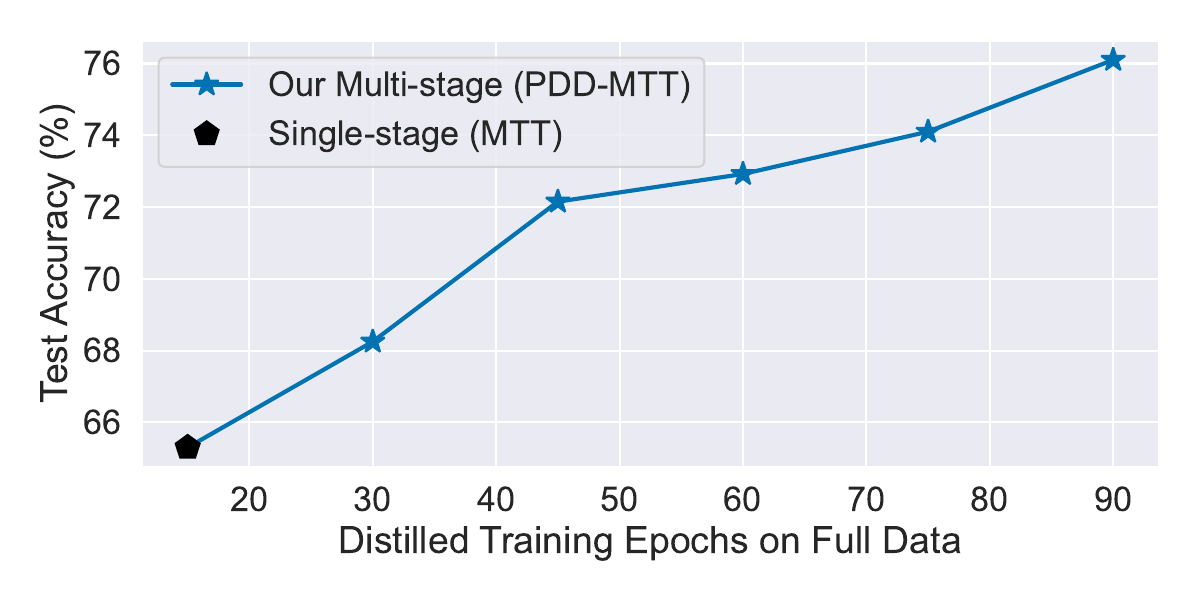}
    \vspace{-2em}
    \caption{Our proposed multi-stage dataset distillation framework \alg improves the state-of-the-art algorithms by distilling longer training dynamics on full data.}
    \vspace{-1em}
    \label{fig:teaser}
\end{wrapfigure}

Dataset distillation aims to generate a very small number of synthetic examples from a large dataset, which can provide a similar generalization performance to that of training on the full dataset \citep{wang2018dataset,loo2022efficient,nguyen2021datasetmeta,nguyen2021datasetdistillation,zhou2022dataset}. If this can be achieved, it can significantly reduce the costs and memory requirement of training deep network on large datasets.
Therefore, dataset distillation has gained a lot of recent interest and found various applications, ranging from continual learning, neural architecture search to privacy-preserving ML \citep{zhao2021dataset,dong2022privacy}.

Existing dataset distillation methods generate a set of synthetic examples that match the gradient \citep{zhao2021dataset,zhao2021dsa}, Neural Tangent Kernel (NTK) \citep{loo2022efficient,nguyen2021datasetmeta,nguyen2021datasetdistillation,zhou2022dataset}, or weights \citep{kim2022dataset} of a number of randomly initialized models being trained on the original \citep{zhao2021dataset} or synthetic data \citep{zhao2021dsa}. However, as matching the entire training dynamics is intractable, existing methods only match the dynamics of \textit{early training iterations}, as short as the first four epochs \citep{zhao2021dataset}. As training dynamics of deep networks drastically change during the training, a synthetic subset generated based on early training dynamics cannot represent the dynamics of later training phases. Hence, existing dataset distillation methods  
suffer from a substantial performance gap to that of training on the original data \citep{zhao2021dataset,kim2022dataset}.

Recent results on optimization and generalization of neural networks revealed that gradient methods have an inductive bias towards learning simple functions, in particular early in training \citep{kalimeris2019sgd,hu2020surprising,hermann2020shapes,neyshabur2014search,shah2020pitfalls}. 
That is, models trained with (stochastic) gradient methods learn nearly linear functions in the initial training iterations \citep{kalimeris2019sgd,hu2020surprising}. As iterations progress, SGD learns functions of increasing complexity \citep{kalimeris2019sgd}.
This implies that synthetic examples generated based on early training dynamics can only train low-complexity neural networks that perform well on easy examples that are separable by low-complexity models. This limitation is further supported by recent studies \citep{pooladzandi2022adaptive,pmlr-v202-yang23g}, which observed that deep models benefit the most from learning examples of increasing difficulty levels at various training stages and one subset of the training data is not enough to support the entire training.

Building on this observation, to bridge the gap to training on the full data, it is crucial to synthesize examples that can capture the dynamics of later training phases. 
This is, however, very challenging. First, synthesizing examples that match the training dynamics of many randomly initialized networks over longer training intervals has a very high computational cost. Moreover, capturing the complex training dynamics over longer intervals 
requires synthesizing more images, which makes it prohibitively expensive. Finally, even if a larger subset can be generated to match the dynamics of a longer training interval, it is not enough to bridge the gap to training on the full data.

In this work, we address the above challenges by proposing a Progressive Dataset Distillation (\alg) pipeline. We are the first to employ a multi-stage idea that focuses on different phases of training. The key idea behind our method is to generate multiple synthetic subsets that can capture the training dynamics in different phases.
To do so, we synthesize examples that capture training dynamics of the full data in a given training interval. Then, we train the model on the synthesized examples and generate another set of synthetic examples that, together with the previous synthetic sets, capture training dynamics of the full data in the consecutive training interval. 
Importantly, our progressive distillation approach effectively trains neural networks with superior generalization performance without increasing the training time on the synthetic examples.
Figure \ref{fig:teaser} confirms that by distilling dynamics of later training stages on CIFAR-10, \alg effectively improves the performance when training on the distilled data.

Our extensive experiments confirm that our multi-stage distillation approach outperforms existing methods by 
up to $5\%$ on ConvNet and $5\%$ for cross-architecture generalization to ResNet-10 and ResNet-18.
Remarkably, \alg is the first method to enable generating larger synthetic datasets. In doing so, it considerably bridges the gap to training on the full data by achieving 90\% of the full accuracy with only $5\%$ of the full data size on CIFAR-10 and $8\%$ of full data size on CIFAR-100 \citep{krizhevsky2009learning} and provides state-of-the-art accuracy on Tiny-ImageNet~\citep{le2015tiny}.
We also conduct studies showing that: 1) our multi-stage synthesis framework achieves consistent improvement if new intervals are introduced; 2) our framework generates synthetic samples with strong generalization ability across various architectures; 3) the distillation process can be performed on progressively challenging subsets of the full data at each stage, resulting in minimal performance degradation. 
\vspace{-1em}
\section{Related Works}
\vspace{-2mm}
\label{sec:related}

Dataset Distillation (\texttt{DD})~\citep{wang2018dataset} aims to generate a synthetic subset from a large training data that can achieve a similar generalization performance to
that of training on the full dataset, when trained on.
To achieve this, 
\texttt{DD} adopted an
optimization process comprising two nested loops. The inner loop involves training a model using the synthesized data until it reaches convergence, while the outer loop aims to optimize the synthetic data such that the trained model generalizes well on the original dataset. 
More recent studies \citep{loo2022efficient,nguyen2021datasetmeta,nguyen2021datasetdistillation,zhou2022dataset} leverage the same framework but use kernel methods, such as Neural Tangent Kernel (NTK), to approximate the inner optimization in a closed form.
While kernel-based algorithms achieved higher accuracy than \texttt{DD}~\citep{wang2018dataset} on networks that satisfy the infinite-width assumption, they do not work well in practice as the constant NTK assumption does not generally hold. 

Another set of methods relies on gradient matching. In particular, \texttt{DC} \citep{zhao2021dataset} minimizes the distance between the gradients of the synthetic and original data on the network being trained on the synthetic data.
\texttt{DSA}~\citep{zhao2021dsa} improves upon \texttt{DC} by applying 
differentiable siamese augmentations to both the original and synthetic data while matching their training gradients. Incorporating differentiable data augmentation has been adopted by almost all subsequent studies.
Later on, \texttt{IDC}~\citep{kim2022dataset} proposed a multi-formulation framework to generate more augmented examples from the same set of synthetic data, to boost the performance with the same storage budget. The synthetic data is generated by minimizing the distance between the gradients of the synthetic and original data on the network being trained on the full data.

Besides matching the gradients, other methods involve matching the training trajectories of the network parameters \citep{cazenavette2022dataset} or the data distribution \citep{wang2022cafe,zhao2023dataset}. \texttt{MTT}~\citep{cazenavette2022dataset} pre-computes and stores training trajectories of expert networks trained on the original data, and then minimizes the distance between the parameters of the network trained on the synthetic data and the expert networks. \texttt{CAFE}~\citep{wang2022cafe} matches the features between the synthetic and real data in all intermediate layers. To avoid the expensive bi-level optimization, \texttt{DM}~\citep{zhao2021datasetdm} minimizes the distance between feature embeddings of the synthetic and real data based on randomly initialized networks. More recently, \texttt{HuBa}~\citep{liu2022dataset} proposed to distill a dataset into two components, Bases and Hallucination to increase
the representation capability of distilled datasets. \texttt{IT-GAN}~\citep{zhao2022synthesizing} inverted the training samples into latent spaces and further fine-tuned towards a distillation objective, and \texttt{GLaD}~\citep{cazenavette2023generalizing} used generative adversarial networks as a prior to help the cross-architecture generalization. 

Existing works generate a set of synthetic examples that match the dynamics of neural networks during early-training stage or at multiple random initializations. 
In contrast, we show that progressively generating multiple synthetic subsets to match the training dynamics in different stages of training yields superior performance.

\vspace{-1em}
\section{Problem Formulation and Preliminary}
\vspace{-0.5em}

Given a large dataset $\T = \{(\x_i, y_i)\}$ which consists of $|\T|$ samples from $C$ classes, dataset distillation aims to learn a synthetic set $\S = \{(\s_i, y_i)\}$ with $|\S|$ synthetic samples so that the deep neural networks can be trained on $\S$ and achieve a comparable generalization performance to those trained on $\T$. Formally,
\begin{equation}
    \E_{\x\sim P(\D)}[\L(\phi_{\thet^{\T}}(\x),y)]\simeq \E_{x\sim P(\D)}[\L(\phi_{\thet^{\S}}(\x),y)],
\end{equation}
where $P(\D)$ is the real data distribution, $\phi_{\thet^{\T}}(.)$ and $\phi_{\thet^{\S}}(.)$ are models trained on $\T$ and $\S$ respectively. $\L(.,.)$ is the
loss function, e.g., cross-entropy loss.

State-of-the-art dataset distillation methods condense the real dataset into a small synthetic set by 
matching the gradient of full data along the synthetic or real trajectory.
This can be expressed as follows:
\begin{equation}\label{eq:grad_match}
    \argmin_{\S}\E_{\thet_0\sim P_{\thet_0}}[\sum_{t=0}^{T-1} D(\nabla_{\thet} \L^{\S}(\thet_t),\nabla_{\thet}\L^{\T}(\thet_t))],
\end{equation}
where $\thet_t$ is the model parameters, and $D$ computes distance between the gradients. 
\texttt{DC}~\citep{zhao2021dataset}, and \texttt{DSA}~\citep{zhao2021dsa} update $\thet_t$ by minimizing the loss $\L^{\S} (\thet_t)$ on the synthetic data. On the other hand, \texttt{IDC}~\citep{kim2022dataset} showed that updating $\thet_t$ by minimizing the loss $\L^{\T} (\thet_t)$ on the full data yields superior performance.
Matching the gradient of the augmented version of the training and synthetic examples further improves the performance \citep{zhao2021dataset,kim2022dataset}.

Alternatively, \texttt{MTT} \citep{cazenavette2022dataset} trains two models on synthetic and real data and matches weight trajectories $\thet_{t+N}$ of length $N$ when training the model on synthetic data $\S$ with weight trajectories $\thet^*_{t+M}$ of length $M\gg N$ when training the model on real data $\T$:
\begin{equation}\label{eq:weight_match}
    \argmin_{\S}\frac{\|\thet_{t+N}-\thet^*_{t+M}\|_2^2}{\|\thet_{t}-\thet^*_{t+M}\|_2^2}.
\end{equation}

Existing dataset distillation methods synthesize examples based on the gradients or weights of the models during the initial training epochs \citep{cazenavette2022dataset,kim2022dataset}, or match outputs of multiple randomly initialized models \citep{zhao2021datasetdm}. 
The most successful methods, synthesize examples that capture the training dynamics of models trained on \textit{full data} \citep{kim2022dataset,zhao2021dataset}.
However, they only capture the \textit{early} training dynamics.
For example, \texttt{IDC}~\citep{zhao2021dataset} and \texttt{MTT}~\citep{kim2022dataset} synthesize examples by matching gradients and weights of the first 4 and 15 epochs of a 200 training pipeline respectively, when distilling CIFAR-10 and CIFAR-100.  This is because matching weights or gradients over longer intervals 
becomes computationally difficult and does not yield high-quality synthetic data. This introduces a performance gap to that of training on the original data.

\begin{figure}
    \centering  \includegraphics[width=0.8\textwidth]{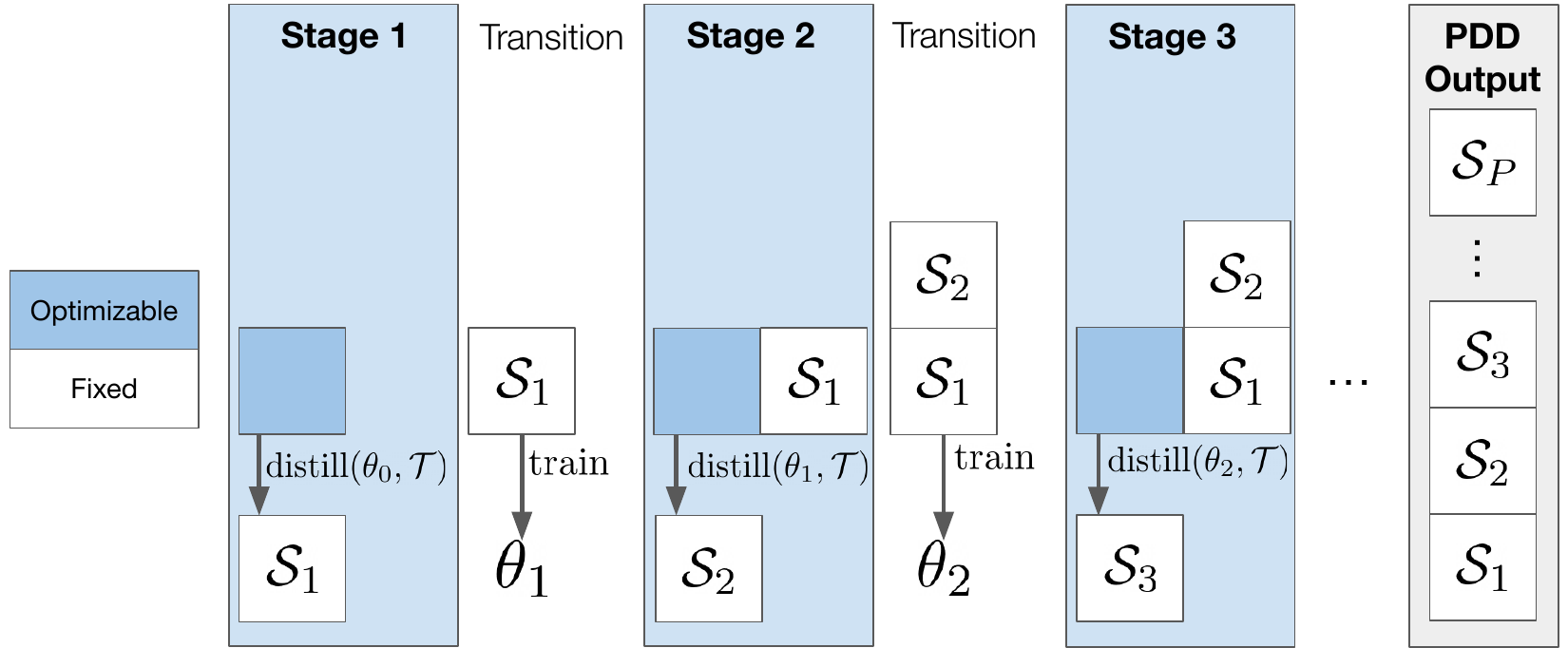}
    \caption{An illustration of the proposed Progressive Dataset Distillation (\alg) framework. It consists of multiple distillation stages and transitions in between. In each distillation stage, we distill a new set of images conditioned on images synthesized in the previous stages. In transitions, we train models on all synthesized images so far, as the starting weights for the next distillation stage to capture longer training dynamics. Our framework can be applied to any dataset distillation algorithm. }
    \label{fig:diagram}
    \vspace{-1em}
\end{figure}

\section{Progressive Dataset Distillation (PDD)}
Next, we introduce our Progressive Dataset Distillation (\alg) framework to generate synthetic images that match the training dynamics of different stages of training.

\begin{wrapfigure}{R}{0.52\textwidth}
    \begin{minipage}{0.52\textwidth}
    \vspace{-7mm}
\begin{algorithm}[H]
\caption{Progressive Dataset Distillation (\alg)}
\label{alg:ours}
\begin{algorithmic}
\STATE \textbf{Input:} A dataset distillation algorithm $\mathcal{A}$, full training set $\mathcal{T}$
\STATE \textbf{Output:} Model trained on a series of synthetic datasets: $\mathcal{S}_1, \mathcal{S}_2, \dots, \mathcal{S}_N$
\STATE{\textbf{Generating synthetic subsets: PDD}}
\STATE $\mathcal{S}_0 \leftarrow \emptyset$
\STATE Initialize $\thet_0$ randomly 
\FOR{$i=1,2,\dots,P$}
\STATE $\S_i=\A(\thet_{i},\T|\cup_{j=1}^{i-1}\S_j)$ 
\STATE $\thet_i=\argmin_{\thet} \L(\thet,\cup_{j=1}^{i-1}\S_j,\thet_{i-1})$
\ENDFOR
\STATE \textbf{Evaluation: Training on the PDD subsets}
\STATE Initialize $\thet_0$ randomly 
\FOR{$i=1,2,\cdots,P$}
\STATE $\thet_i=\argmin_{\thet} \L(\thet,\cup_{j=1}^{i-1}\S_j,\thet_{i-1})$
\ENDFOR
\end{algorithmic}
\end{algorithm}
\end{minipage}
\vspace{-3mm}
\end{wrapfigure}

\subsection{Distilling Multiple Training Stages}
To capture the learning dynamics of different stages of training, our key idea, shown in \cref{fig:diagram}, is to generate a \textit{sequence} of small synthetic datasets $\S_1,\S_2,\cdots,\S_P$, such that each synthetic dataset $\S_i$ captures dynamics of training on the full data in a different stage of training. Then at test time, when the model is trained on the synthetic images, we can train the model on different subsets to mimic different stages of training on the full data. 

However, naively dividing the full training pipeline into $P$ intervals and generating a subset based on the training dynamics of each interval does not yield a satisfactory performance,
due to the following reasons.
First, generating different synthetic subsets independently results in capturing redundant information in the subsets and does not improve the performance.
Second, since the synthetic subsets are small, at test time when the model is trained on the synthetic images, minimizing the loss on subset $\S_{i+1}$ results in forgetting the previously learned information from subsets $\S_1,\cdots \S_i$.
Finally, even if forgetting can be prevented, 
transitioning from training on $\S_i$ to training on $\S_{i+1}$ at test time changes the training loss and interrupts the optimization pipeline. This does not allow the model to learn well from multiple synthetic subsets. 

To address the above issues, we synthesize each subset $\S_{i}$ based on the dynamics of training on the full data at stage $i$, \textit{conditioned} on the previous subsets $\S_1,\S_2,\cdots,\S_{i}$. 
That is, we generate $\S_{i}$ such that $\S_1\cup\S_2\cup \cdots\cup\S_{i}$ captures the training dynamics at stage $i$. Note that we only synthesize $\S_i$ at interval $T_i$ while keeping $\S_1,\S_2,\cdots,\S_{i-1}$ fixed. This prevents capturing redundant information in subset $\S_i$ that are already captured by previous subsets $\S_1,\S_2,\cdots,\S_{i-1}$. 
Next, to address the discontinuity in training on multiple subsets, we synthesize every subset $\S_i$ based on the training dynamics of full data starting from parameters $\thet_i$, where training on $\S_1\cup\S_2\cup \cdots\cup\S_{i-1}$ is finished. This allows smooth transitioning between different subsets when training on the synthetic data.
Finally, at test time when the model is trained on the synthetic subsets, to prevent forgetting the information learned from the previous subsets, we first train the model on $\S_1$, then $\S_1\cup\S_2$, and keep training on the union of the previous subsets in addition to the new one $\S_1\cup\S_2\cup \cdots\cup\S_i$.

We summarize our pipeline in \Cref{alg:ours}. Formally, for $i=1,\cdots, P$, we generate a synthetic subset $\S_i$ as follows:
\begin{equation}
    \S_i=\A(\thet_{i},\T|\cup_{j=1}^{i-1}\S_j) \quad\quad \text{s.t.}  \quad\quad  \thet_i=\argmin_{\thet} \L(\thet,\cup_{j=1}^{i-1}\S_j,\thet_{i-1}),
\end{equation}
where $\L(\thet,\S,\thet_{i-1})$ is the loss of the model trained on data $\S$ starting from $\thet_{i-1}$.
$\A$ can be any dataset distillation method, such as \texttt{DC} \citep{zhao2021dataset}, \texttt{DSA} \citep{zhao2021dsa}, \texttt{IDC} \citep{zhao2021dataset}, and \texttt{MTT} \citep{kim2022dataset}, described in Eq. \eqref{eq:grad_match} and \eqref{eq:weight_match}.\looseness=-1

\textbf{Distillation and training costs.} Note that conditioning the distillation on previous subsets does not increase the cost of synthesizing a new subset, as we generate the same number of synthetic images at every interval. On the other hand, at test time, we train on similar number of images in total with multiple stages. This is because instead of training on $k=|\S|$ synthetic examples during the entire training, \texttt{PDD} with $m$ intervals first trains the model on $k/m$ synthetic images. Then, it trains the model on $2k/m$ synthetic images and keeps increasing the number of training examples until it trains on $k$ examples at the final interval.

\vspace{-1mm}
\subsection{Discarding Easier-to-learn Examples at Later Stages}
\vspace{-1mm}
As training progress, \alg generates synthetic examples that enable the network to learn higher complexity functions. This implies that at later stages, we can safely discard the examples that are learned early in training with lower-complexity functions from the distillation pipeline.
To calculate the learning difficulty of examples, we use the forgetting score \citep{toneva2018an} defined as the number of times the prediction of every example changes from correct to wrong during the training. 
Examples with higher forgetting scores are learned later during the training with higher complexity functions. On the other hand, examples that have a very low forgetting score are those that can be classified by lower complexity functions, early in training. At every distillation stage, we drop examples with low forgetting scores and focus the distillation on examples with increasing levels of difficulty, measure by forgetting score. This improves the efficiency of \alg without harming the performance, as we will confirm experimentally.

Next, we will show experimentally that \texttt{PDD} effectively trains higher-quality neural networks with superior generalization performance without increasing the training time on the synthetic examples.

\vspace{-2mm}
\section{Experiments}
\vspace{-0.5em}

In this section, we assess the classification performance of neural networks trained on synthetic images generated by our framework. In addition to evaluating on the architecture used for distillation, we also investigate the transferability of the distilled images to larger models with different architectures. We further show with ablation studies that {\alg} trains models with increasing classification accuracy when we increase the number of intervals, and confirm the importance of 
conditioning and transitions.
\begin{table}[t]
    \centering
    \caption{Test accuracy of ConvNets on CIFAR-10/100 and Tiny-ImageNet, trained on synthetic samples generated by various models with different numbers of images per class (IPC). Our algorithm (\alg) improves upon baseline methods through its multi-stage distillation pipeline, narrowing the performance gap relative to training on the full dataset. \alg results are reported for 5 stages.
    }
    \label{tab:syn_dataset}
    \vspace{-3mm}\resizebox{1\linewidth}{!}{
    \begin{tabular}{cc|cc|cc|cc}
       \toprule & Dataset & \multicolumn{2}{c}{CIFAR-10} & \multicolumn{2}{c}{CIFAR-100} & \multicolumn{2}{c}{Tiny-ImageNet} \\
       \midrule & IPC & 10 & 50 & 10 & 50 & 10 & 50 \\
       \midrule \multirow{4}{*}{Selection} & Random  & $26.0\pm1.2$ & $43.4\pm1.0$ & $14.6\pm0.5$  &  $30.0\pm0.4$ & $5.0\pm0.2$ & $15.0\pm0.4$\\
       & Herding  & $31.6\pm0.7$ & $40.4\pm0.6$ & $17.3\pm0.3$  & $33.7\pm0.5$ & $6.3\pm0.2$ & $16.7\pm0.3$ \\
       & K-Center & $14.7\pm0.9$ & $27.0\pm1.4$ & $7.1\pm0.2$  & - & - & - \\
       & Forgetting & $23.3\pm1.0$ & $23.3\pm1.1$ & $15.1\pm0.2$   & $30.5\pm0.3$ & $5.1\pm0.2$ & $15.0\pm 0.3$ \\
       \midrule
       \multirow{9}{*}{Distillation} & \texttt{DC} & $44.9\pm0.5$ & $53.9\pm0.5$ & $25.2\pm0.3$  & - & - & - \\
       & \texttt{DSA} & $52.1\pm0.5$ & $60.6\pm0.5$ & $32.3\pm0.3$   & $42.8\pm0.4$ & - & - \\
       
        & \texttt{CAFE} & $46.3\pm0.6$ & $55.5\pm0.6$ & $27.8\pm0.3$   & $37.9\pm0.3$ & - & - \\
       & \texttt{CAFE} + \texttt{DSA} & $50.9\pm0.5$ & $62.3\pm0.4$ & $31.5\pm0.2$   & $42.9\pm0.2$ & - & - \\
       & \texttt{DM} & $48.9\pm0.6$ & $63.0\pm0.4$  & $29.7\pm0.3$  & $43.6\pm0.4$  & $12.9\pm0.4$ & $24.1\pm0.3$ \\

        & \texttt{MTT} & $65.3\pm0.7$ & $71.9\pm0.2$ & $39.6\pm0.3$ & $47.7\pm0.2$ & $23.2\pm0.3$ & $28.0\pm0.3$ \\
        \rowcolor{lightgray!40} \cellcolor{white} &  \texttt{\textbf{PDD}+MTT} & $\mathbf{66.9\pm0.4}$ & $\mathbf{74.2\pm0.5}$ & $\mathbf{43.1\pm 0.7}$ & $\mathbf{52.0\pm0.5}$ & $\mathbf{27.3\pm0.5}$ & $\mathbf{29.2\pm0.6}$ \\
        & \texttt{IDC}  & $67.5\pm0.5$ & $74.5\pm0.1$  & $45.1\pm0.3$ & $52.5\pm0.4$ & - & - \\

        \rowcolor{lightgray!40}  \cellcolor{white} &  \texttt{\textbf{PDD}+IDC} & $\mathbf{67.9\pm0.2}$ & $\mathbf{76.5\pm0.4}$ & $\mathbf{45.8\pm0.5}$ & $\mathbf{53.1\pm0.4}$ & - & -   \\
      \midrule
      \multicolumn{2}{c|}{Full Data} & \multicolumn{2}{c|}{88.1} & \multicolumn{2}{c|}{56.2} & \multicolumn{2}{c}{37.6} \\
        \bottomrule
    \end{tabular}}%
    \label{tab:syn_dataset_break}
    
\vspace{-3mm}
\end{table}

\begin{table}[ht]

\end{table}

\vspace{-3mm}
\subsection{Experimental Settings}

\textbf{Datasets.}  We conduct our experiments on three standard datasets: CIFAR-10, CIFAR-100~\citep{krizhevsky2009learning} and Tiny-ImageNet~\citep{le2015tiny}. CIFAR-10 and CIFAR-100 consist of $50,000$ training images, with $10$ and $100$ classes, respectively. The image size for CIFAR is $32\times 32$. Tiny-ImageNet contains $100,000$ training images from $200$ categories, with the image size of $64 \times 64$.  

\textbf{Baselines. } We consider both data selection and distillation algorithms as baselines, including random selection, Herding \citep{welling2009herding}, K-center~\citep{farahani2009facility}, and Forgetting~\citep{toneva2018an} for selection and \texttt{DC}~\citep{zhao2021dataset}, \texttt{DSA}~\citep{zhao2021dsa}, \texttt{DM}~\citep{zhao2021datasetdm}, \texttt{CAFE}~\citep{wang2022cafe}, \texttt{IDC}~\citep{kim2022dataset}, and \texttt{MTT}~\citep{cazenavette2022dataset} for distillation. Herding~\citep{welling2009herding} greedily selects samples to approximate the mean of the entire dataset; Forgetting score~\citep{toneva2018an} keeps track of how many times a training sample is learned and forgotten during the training and keeps examples with the highest forgetting score; 
K-Center~\citep{farahani2009facility} selects the samples to minimize the maximum distance between a data point and its center. Distillation baselines are introduced in \cref{sec:related}.

\textbf{Architectures.} Our experimental settings follow that of \citet{cazenavette2022dataset}: we employ a ConvNet for distillation, with three convolutional blocks for CIFAR-10 and CIFAR-100 and four convolutional blocks for Tiny-ImageNet, each containing a 128-kernel convolutional layer, an instance normalization layer~\citep{ulyanov2016instance}, a ReLU activation function~\citep{nair2010rectified} and an average pooling layer. 
We include ResNet-18 and ResNet-10~\citep{he2016deep} to assess the transferability of the synthetic images to other architectures.

\textbf{Distillation Settings.}
We adopt two representative baseline methods on which we apply our framework: \texttt{IDC} and \texttt{MTT}, which are widely used state-of-the-art dataset distillation methods. 
During the matching process, we adopt the optimal hyper-parameter reported in the original paper of each dataset distillation method in each stage of \alg without further tuning.
We report the number of images \alg distills at each stage 
and also report the number of synthetic sets $P$ in our results to enable a comparison between \alg and the baselines. 
Note that the number of synthetic sets has a monotonic effect on the models' testing accuracies.

\textbf{Evaluation.} Once the synthetic subsets have been constructed for each dataset, they are used to train randomly initialized networks from scratch, followed by evaluation on their corresponding testing sets. For \alg, we sequentially train models after each interval on all synthetic samples that have already been generated up to the current interval. For each experiment, we report the mean and the standard deviation of the testing accuracy of $5$ trained networks. 
To train networks from scratch at evaluation time, we use the SGD optimizer with a momentum of $0.9$ and a weight decay of $5\times 10^{-4}$. For \texttt{IDC}, the learning rate is set to be $0.01$. For \texttt{MTT}, the learning rate is simultaneously optimized with the synthetic images. During the evaluation time, we follow the augmentation strategies of each method to train networks from scratch.

\subsection{Evaluating Distilled Datasets}

\textbf{Setup.} We demonstrate the effectiveness of the proposed multi-stage distillation by applying $\alg$ to \texttt{MTT} and \texttt{IDC} to distill CIFAR-10/100 and Tiny-ImageNet. 

\cref{tab:syn_dataset_break} compares \alg with state-of-the-art baselines for different values of Images Per Class (IPC) distilled in 5 stages. We specify baselines' IPC and \alg's IPC to be $10$ and $50$ for all the benchmarks. 
For Tiny-ImageNet, we only conduct experiments with \texttt{MTT} as \texttt{IDC}'s distillation time is prohibitively expensive 
in this higher resolution. 
Based on the default settings, single-stage \texttt{IDC} distills 4 epochs of training on the real images; \texttt{MTT} distills $15$ epochs for CIFAR-10, $20$ for CIFAR-100, and $40$ for Tiny-ImageNet.

\textbf{Comparing to Single Stage Distillation. } We see that \alg consistently improves the performance across all data selection and distillation baselines with the same IPCs, especially when we distill longer training dynamics (i.e., $15$ epochs with \texttt{MTT}) on the real images in each stage. 
Specifically, \alg + \texttt{MTT} outperforms \texttt{MTT} by significant margins of $1.6\%/2.3\%$ on CIFAR-10 IPC-10/50, $3.5\%/4.3\%$ on CIFAR-100, and $ 4.1\%/1.2\%$ on Tiny-ImageNet. After applying \alg to \texttt{IDC}, we witness an substantial improvement on performance across different datasets: $0.4\%/2.0\%$ on CIFAR-10 IPC-10/50, respectively, and $0.7\%/0.6\%$ on CIFAR-100 IPC-10/50, respectively.

\begin{figure}[t]
     \centering
     \begin{subfigure}[b]{0.32\textwidth}
         \centering
         \includegraphics[width=\textwidth]{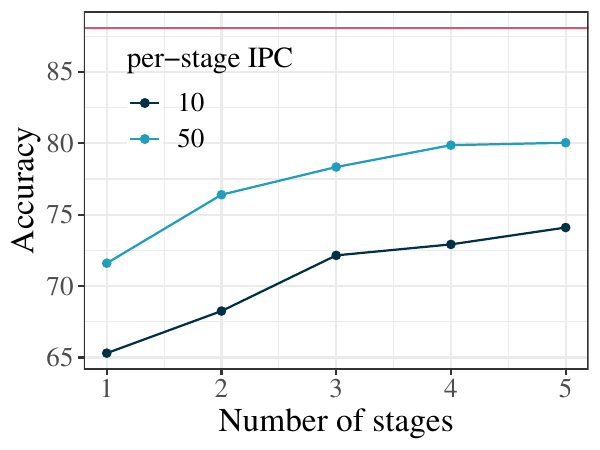}
         \caption{CIFAR-10}
         \label{fig:cifar10_trend}
     \end{subfigure}
     \begin{subfigure}[b]{0.32\textwidth}
         \centering
         \includegraphics[width=\textwidth]{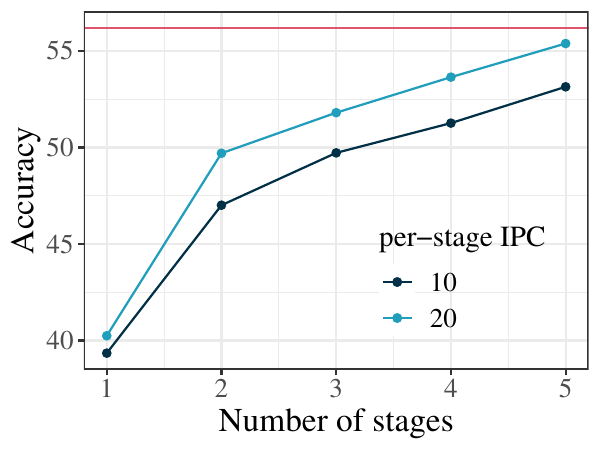}
         \caption{CIFAR-100}
         \label{fig:cifar100_trend}
     \end{subfigure}
     \begin{subfigure}[b]{0.32\textwidth}
         \centering
         \includegraphics[width=\textwidth]{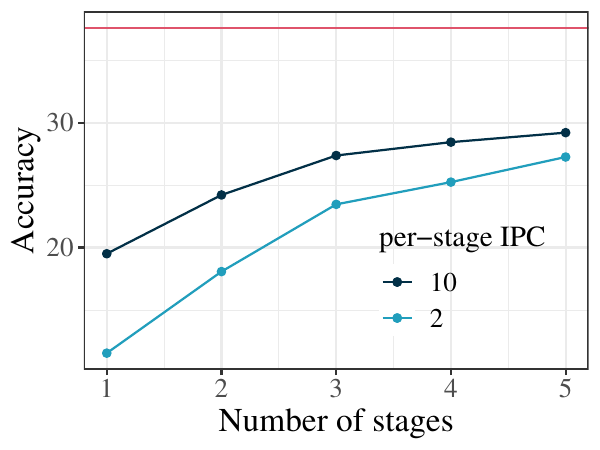}
         \caption{Tiny-ImageNet}
         \label{fig:tiny-imagenet}
     \end{subfigure}
     \vspace{-2mm}
     \caption{ConvNets' test accuracy on CIFAR-10, CIFAR-100 and Tiny-ImageNet after training on samples distilled by \alg + \texttt{MTT} with multiple stages with larger per-stage IPCs. Left: performance on CIFAR-10; Middle: performance on CIFAR-100; Right: performance on Tiny-ImageNet. The red lines indicate the performance of training on full data on CIFAR-10, CIFAR-100 an Tiny-ImageNet, respectively.}
     \vspace{-1em}
\end{figure}

\textbf{Scaling up synthetic datasets: towards bridging the gap  to training on the full data.
} 
In Figure~\ref{fig:cifar10_trend} and \ref{fig:cifar100_trend}, we extend our experiments with \texttt{MTT} by maintaining a constant per-stage IPC while progressively increasing the number of stages. 
This setting 
enables scaling of synthesis process to generate larger total IPC, because the images generated in earlier stages 
are employed in subsequent stages. We conduct these experiments on CIFAR-10 and CIFAR-100, respectively, and set the per-stage IPC to $10/50$ for CIFAR-10, $10/20$ for CIFAR-100, and $2/10$ for Tiny-ImageNet. Remarkably, \alg considerably bridges the gap to training on the full dataset by achieving 90\% of the full accuracy with only $5\%$ of the full data size on CIFAR-10 (which means that IPC $=250$) and $10\%$ of full data size on CIFAR-100 (which means that IPC $=50$). Notably, for CIFAR-100, we utilize $20\%$ of the complete dataset, resulting in an IPC value of $100$, yet achieve a comparable performance. On Tiny-ImageNet, applying \texttt{PDD} with \texttt{MTT} could also reach $80\%$ of the performance obtained by training on the full data after distilling $50$ images per class. 

\begin{wraptable}{r}{0.5\linewidth}
    \centering
    \vspace{-1.5em}
    \caption{Performance on other architectures of networks with synthetic datasets generated on ConvNets by baselines versus \alg+ baselines.}
    \resizebox{!}{0.25\linewidth}{
    \begin{tabular}{c|ccc}
    \toprule
        (Total) IPC &  Method & ResNet-10 & ResNet-18 \\
        \midrule \multirow{4}{*}{10} & \texttt{IDC} & \cellcolor{white} $63.0\pm0.6$ & \cellcolor{white} $63.6\pm0.4$ \\
         & \cellcolor{lightgray!40} \texttt{\textbf{PDD}+IDC} & \cellcolor{lightgray!40} $\mathbf{63.2\pm1.4}$ & \cellcolor{lightgray!40} $\mathbf{63.9\pm0.6}$\\
        \cdashline{2-4} & \texttt{MTT} & \cellcolor{white} $46.4\pm0.4$ & \cellcolor{white} $45.2\pm0.3$ \\
         & \cellcolor{lightgray!40} \texttt{\textbf{PDD}+MTT} &  \cellcolor{lightgray!40} $\mathbf{47.1\pm0.3}$ & \cellcolor{lightgray!40} $\mathbf{46.0\pm0.4}$\\
        \midrule \multirow{4}{*}{50} & \texttt{IDC} & \cellcolor{white} $70.7\pm0.6$ & \cellcolor{white} $69.8\pm0.3$ \\
         & \cellcolor{lightgray!40} \texttt{\textbf{PDD}+IDC} & \cellcolor{lightgray!40} $\mathbf{72.4\pm0.4}$ & \cellcolor{lightgray!40} $\mathbf{71.6\pm0.3}$ \\
        \cdashline{2-4} & \texttt{MTT} & \cellcolor{white} $63.1\pm0.4$ & \cellcolor{white} $62.6\pm0.4$\\
         & \cellcolor{lightgray!40} \texttt{\textbf{PDD}+MTT} & \cellcolor{lightgray!40} $\mathbf{64.6\pm0.9}$  & \cellcolor{lightgray!40} $\mathbf{63.5\pm0.5}$\\
         \bottomrule
     \end{tabular}}
    \label{tab:transfer}
    \vspace{-1.5em}
\end{wraptable}
\vspace{-3mm}

\vspace{-0.5em}
\subsection{Cross-Architecture Generalization}
\vspace{-0.5em}

Next, we evaluate the generalization performance of \alg on architectures that are different from the one we used to distill CIFAR-10. Following the settings of cross-architecture experiments in the original papers, we use batch normalization layers when evaluating on \texttt{IDC}, and use instance normalization layers for \texttt{MTT}. We follow the same evaluation pipeline for each baseline method to acquire and present the test accuracy in Table~\ref{tab:transfer}. 

We can see that the images distilled by using \alg improves other architectures' performance ($1.7\%/1.5\%$ on ResNet-10 and $1.8\%/0.9\%$ on ResNet-18) when using IPC $=50$, and show considerable improvement ($0.2\%/0.7\%$ on ResNet-10 and $0.3\%/0.8\%$ on ResNet-18) compared to using the single-stage \texttt{MTT} and \texttt{IDC} when the total IPC is $10$.  
These results indicate our distilled images from multiple stages are robust to changes in network architectures.

\vspace{-1mm}
\subsection{Ablation Studies}
\label{sec:ablations}

\textbf{Effect of Progressive Training. }  
When training a model on the $P$ synthetic subsets, \alg progressively trains on the union of the first $i$ synthetic sets, for $i=1,\cdots, P$. To demonstrate the effectiveness of this progressive way of training,  we explore multiple choices of training pipelines with the \alg generated synthetic sets: (1) 
\textit{Union}: we train on the union of the synthetic sets generated in all $P$ stages, i.e., $\cup_{j=1}^P \S_j$; 
(2) 
\textit{Sequential}: we train on different $\S_i$ in the order they are generated;
(3) \textit{Progressive}: 
we progressively train on union of the first $i$ synthetic sets, i.e., $\cup_{j=1}^i \S_j$.

\begin{wraptable}{r}{0.46\linewidth}
    \centering
    \caption{Effect of training on \alg distilled subsets. Testing accuracy on CIFAR-10 after being trained on 10 IPC per stage distilled by \alg + different base methods. 
    In `Training' column, U, S, P correspond to training on $\cup_{j=1}^P\S_j$, or $\S_i$, or $\cup_{j=1}^i \S_j$, at stage $i$, respectively. }
    \resizebox{!}{0.46\linewidth}{
    \begin{tabular}{c|c|c|c}
    \toprule
        \multirow{2}{*}{$P$} & \multirow{2}{*}{Training} & \multicolumn{2}{c}{Test Accuracy}  \\

        \cmidrule{3-4}
        & & \texttt{MTT} + \alg & \texttt{IDC} + \alg \\
        \midrule 
        1 & - & $65.3\pm0.7$ & $67.5\pm0.5$
        \\
        \midrule
        \multirow{3}{*}{2} & U & $60.4\pm0.6$  & $71.1\pm0.2$ \\
        & S & $64.1\pm1.0$ & $68.5\pm0.1$ \\
        \rowcolor{lightgray!40} \cellcolor{white} &  P & $\textbf{68.7}\pm0.8$ & $\textbf{71.4}\pm0.2$ \\
        \midrule
        \multirow{3}{*}{3} & U & $65.4\pm0.7$ & $\textbf{74.2}\pm0.4$ \\
        & S & $67.4\pm1.1$ & $68.2\pm0.7$ \\
        \rowcolor{lightgray!40} \cellcolor{white} &  P & $\textbf{71.5}\pm0.4$ & $74.0\pm0.3$ \\
        \midrule
        \multirow{3}{*}{4} & U & $63.2\pm0.7$ & $75.4\pm0.3$ \\
        & S & $66.0\pm0.9$ & $69.9\pm0.5$\\
        \rowcolor{lightgray!40} \cellcolor{white} &  P & $\textbf{73.1}\pm0.6 $ & $75.4\pm0.1$ \\   
        \midrule
        \multirow{3}{*}{5} & U & $65.9\pm0.4$ & $76.2\pm0.6$  \\
        & S & $67.4\pm0.8 $ & $69.9\pm0.5$ \\
        \rowcolor{lightgray!40} \cellcolor{white} &  P & $\textbf{74.2}\pm0.5$ & $\textbf{76.5}\pm0.2$\\
        \bottomrule
    \end{tabular}
    }
    \vspace{-4mm}
    \label{tab:evaluation_new}
\end{wraptable}

\cref{tab:evaluation_new} compares the above training methods when evaluating the synthetic sets \alg distilled for CIFAR-10 with a fixed per-stage IPC $=10$ and different numbers of stages $P$. For all the base distillation algorithms, namely \texttt{MTT} and \texttt{IDC}, progressive training is consistently better than union and outperforms sequential training with a large margin in particular for larger $P$. This confirms the necessity of progressive training to prevent forgetting the previously learned information.
Note that \alg+ \texttt{MTT} performs poorly with the union pipeline because \texttt{MTT} learns the learning rate for each set of synthetic images, so a single learning rate is not suitable for training on the union.

\textbf{Importance of transitions and conditioning. } There are two key designs in \alg that are essential for the success of multi-stage dataset distillation: (1) \underline{\textit{transition}} between stages by generating a new synthetic subset based on the training trajectory starting from the point where training on the union of the previous synthetic subsets is finished; and (2) \underline{\textit{conditioning}} on synthetic images distilled in earlier stages when generating a new synthetic set for the current training stage. 
In \cref{tab:independent}, we show both components are crucial by comparing the test accuracy of ConvNet after being trained on the \alg distilled datasets with both or without one of the two designs. For \alg+ \texttt{MTT} and both variants, we fix the number of images per class to distill in each stage to be $10$. We observe a decreased performance of \alg when it distills images for each training stage independent of the previous stages, and the difference is more significant when we distill longer training intervals with more stages. \looseness=-1

\begin{table}[ht]
\begin{minipage}{0.58\linewidth}
\centering
            \caption{ConvNet's performance on CIFAR-10 with different synthesis modes (\textit{i.e.}, w/o transition and w/o conditioning) using \alg+ \texttt{MTT}. }
            \resizebox{1\textwidth}{!}{
            \begin{tabular}{c|c|c|c}
                \toprule $P$ & w/o transition & w/o conditioning & \alg \\
                \midrule $1$ & $65.3\pm0.7$ & $65.3\pm0.7$ & $65.3\pm0.7$  \\
                $2$ & $66.0\pm0.6$ & $67.9\pm0.5$ & $68.7\pm0.8$\\
                $3$ & $66.3\pm0.4$ & $69.8\pm0.9$ & $71.5\pm0.4$\\
                $4$ & $65.6\pm0.5$ & $71.4\pm0.5$ & $73.1\pm0.6$\\
                $5$ & $63.6\pm0.7$ & $71.9\pm0.7$ & $74.2\pm0.5$\\
                \bottomrule
            \end{tabular}
            }
            \label{tab:independent}
\end{minipage}\hfill%
\begin{minipage}{0.4\linewidth}
\centering
    \caption{Models' testing accuracy on CIFAR-10. \alg with different numbers of stages ($P$) and per-stage IPC. }
    \resizebox{1\linewidth}{!}{
    \begin{tabular}{c|c|c}
    \toprule
        $P$ & per-stage IPC  & Accuracy \\
        \midrule
        $1$ & $10$ & $65.3\pm0.7$ \\
        $2$ & $5$ & $65.5\pm0.9$ \\
        $5$ & $2$ & $66.9\pm0.4$ \\
        $10$ & $1$ & $64.4\pm0.6$ \\
        \bottomrule
    \end{tabular}}
    \label{tab:assignment}
\end{minipage}
\end{table}

\textbf{Distilling more training stages vs more images per stage. }
Given a fixed total number of images per class, 
we can distill longer training dynamics by having more stages, or choose to distill more images in each stage to capture the dynamics better. To understand which of the above two strategies leads to better performance, we study four different combinations of the number of stages and per-stage IPC, and record the models' test accuracy in \cref{tab:assignment}. We observe that establishing more stages can generally improve the results, as long as per-stage IPC is not too small (IPC $=1$ per stage leads to degraded performance). In particular, with $10$ as a fixed number of images in total, best result corresponds to $P=5$ and per-stage IPC $=2$.

\textbf{Discarding easy-to-learn examples at later stages. }
Next, we confirm that easier-to-learn examples can be dropped from the distillation pipeline in later intervals. To do so, we use the forgetting score~\citep{toneva2018an} defined as the number of times the prediction of every example changes from being correctly classified to incorrectly classified during the training. Examples with higher forgetting scores are more difficult to learn for the network and are learned later during the training \citep{toneva2018an}. 

\begin{wraptable}{r}{0.5\linewidth}
        \centering
            \caption{ConvNet's performance on CIFAR-10 trained on synthetic set with $10$ images per class using \texttt{MTT} with \alg by distilling from easy to difficult samples. In $i$-th stage we select samples with forgetting score within $[3(i-1), 3i)$.  We report the portion of training samples used in each setting. }
            \vspace{-1em}
            \resizebox{0.8\linewidth}{!}{
            \begin{tabular}{c|c|c}
                \toprule
               $P$ & Data Used & Testing Accuracy \\
               \midrule
               \multirow{2}{*}{1} & $36.5\%$ & $65.9\%$ \\
               & $100\%$ & $65.3\%$ \\
               \midrule \multirow{2}{*}{3} & $55.8\%$ & $71.6\%$ \\
               & $100\%$ &  $71.5\%$\\
               \midrule \multirow{2}{*}{5} & $66.4\%$ & $73.7\%$\\
               & $100\%$ & 74.2\% \\
               \bottomrule
            \end{tabular}}
            \vspace{-3mm}
            \label{tab:easy_to_difficult}
\end{wraptable}
We separate training examples into multiple partitions based on their forgetting scores, with an increment of $3$. More specifically, at the $i$-th stage only the examples with a number of forgetting events between $3\times(i-1)$ and $3\times i$. Subsequently, we apply \alg to distill the corresponding partition of data examples at each stage, starting from the partition that contains examples with the lowest forgetting scores and progressing to those with the highest scores. \cref{tab:easy_to_difficult} shows that when \alg explicitly distills examples with increasing learning difficulty at different stages, models trained on the distilled images have comparable test performance as when the distillation is based on the full training set at all stages. This observation not only confirms that \alg naturally creates a curriculum with its synthetic sets but also confirms the possibility of reducing the distillation cost of \alg as the training examples used in each stage can be significantly reduced. 

\vspace{-1mm}
\subsection{Continual Learning}
\vspace{-1mm}
\begin{wraptable}{r}{0.6\linewidth}
    \centering
    \vspace{-2em}
    \caption{Continual learning performance using distilled samples generated by different methods on CIFAR-100.}
    \vspace{-2mm}\resizebox{1\linewidth}{!}{
    \begin{tabular}{c|c|c|c|c|c}
    \toprule
        Methods & Stage1 & Stage 2 & Stage 3 & Stage 4 & Stage 5  \\
        \midrule \texttt{DSA} & $52.5$ & $45.7$ & $40.4$ & $35.0$ & $31.1$\\
        \texttt{Herding} & $48.4$ & $43.3$ & $39.6$ & $36.4$ & $33.1$ \\
        \texttt{MTT} & $55.7$ & $52.1$ & $48.3$ & $43.0$ & $41.2$ \\
        \rowcolor{lightgray!40} \texttt{\textbf{PDD}+MTT} & $\mathbf{61.2}$ & $\mathbf{56.6}$ & $\mathbf{51.5}$ & $\mathbf{48.3}$ & $\mathbf{45.1}$\\

        \bottomrule
    \end{tabular}}
    \vspace{-1em}
    \label{tab:continual}
\end{wraptable}

In this section, we adopt a class incremental setting~\citep{zhao2021dataset, zhao2021dsa} to show that \alg can improve the performance in the application of continual learning. We apply \alg on \texttt{MTT} to distill CIFAR-100 across $5$ phases, in each of which we can only access $20$ classes with $20$ images distilled in total per class. During the evaluation, a model will be trained sequentially on samples available at each stage. Table~\ref{tab:continual} shows the performance using different methods, which demonstrates that \alg+ \texttt{MTT} consistently outperforms \texttt{MTT} at each stage and showcases \alg's ability to improve baselines' performance in the application of continual learning.

\vspace{-3mm}
\subsection{Synthesized Samples Visualization}
\begin{wrapfigure}{r}{0.5\linewidth}
    \centering
    \vspace{-2em}
    \includegraphics[width=1\linewidth]{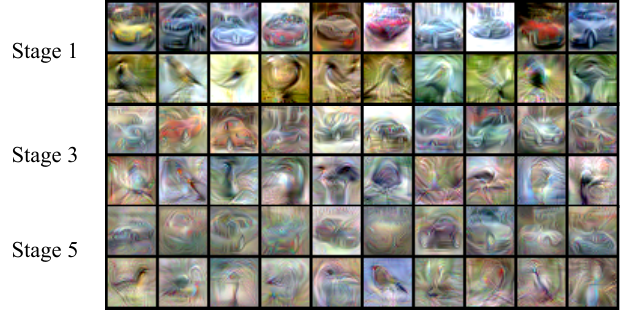}
    \vspace{-1em}
    \caption{Synthesized images of CIFAR-10 using \alg+ \texttt{MTT} from Stage $1$, $3$ and $5$. The images from classes ``automobile'' and ``birds'' at each stage are selected for demonstration. }
    \label{fig:visualization}
    \vspace{-2em}
\end{wrapfigure}

In \cref{fig:visualization}, we provide examples of synthetic samples on CIFAR-10 using \alg+ \texttt{MTT} at different stages. We distill CIFAR-10 in $5$ stages with a per-stage IPC of $10$. From the images we can observe that the synthetic samples at later stages show diversified patterns, demonstrating lower saturation in color and more abstract textures. This evolution of visual patterns indicates a shift in the focus of the distillation process and thus provides an empirical support to our multi-stage design. 
\cref{fig:full_ipc10} shows all the samples from Stage $1$ to $5$ where the transition of distilled patterns on all classes are clearly presented.

\section{Conclusion}

In this work, we proposed a progressive dataset distillation framework, \alg, that generates multiple sets of synthetic samples sequentially, conditioned on the previous ones, to capture dynamics of different training intervals. 
Extensive experiments confirm the effectiveness of \alg in improving the performance of existing dataset distillation methods on various benchmark datasets. 

\clearpage

\bibliography{main}
\bibliographystyle{conference}

\appendix
\renewcommand{\thepage}{A\arabic{page}}  
\renewcommand{\thesection}{A\arabic{section}}   
\renewcommand{\thetable}{A\arabic{table}}   
\renewcommand{\thefigure}{A\arabic{figure}}


\section{Experiment Details}

\subsection{Experiment Settings}
On CIFAR-10, the networks are trained for $\frac{2000}{P + 1}$ epochs at each stage. Consequently, the total iterations taken is $\frac{P(P+1)}{2} \frac{2000}{P + 1} \times \frac{n}{B} = \frac{1000Pn}{B}$, where $B$ is the batch size and $n$ is the number of images newly distilled at each stage. This quantity proves to be adequate in achieving favorable outcomes without inflating the computational burden of network training. Notably, it aligns with utilizing all available images for a training duration of $1000$ epochs. Additionally, it is important to note that augmenting the number of epochs could lead to further enhancements in the test accuracy of the trained networks. For CIFAR-100, the networks undergo training for $500$ epochs during each stage to facilitate improved convergence.

\vspace{-1em}
\section{More Visualization}
\begin{figure}[ht]
    \centering
    \vspace{-1em}
    \includegraphics[width=1\linewidth]{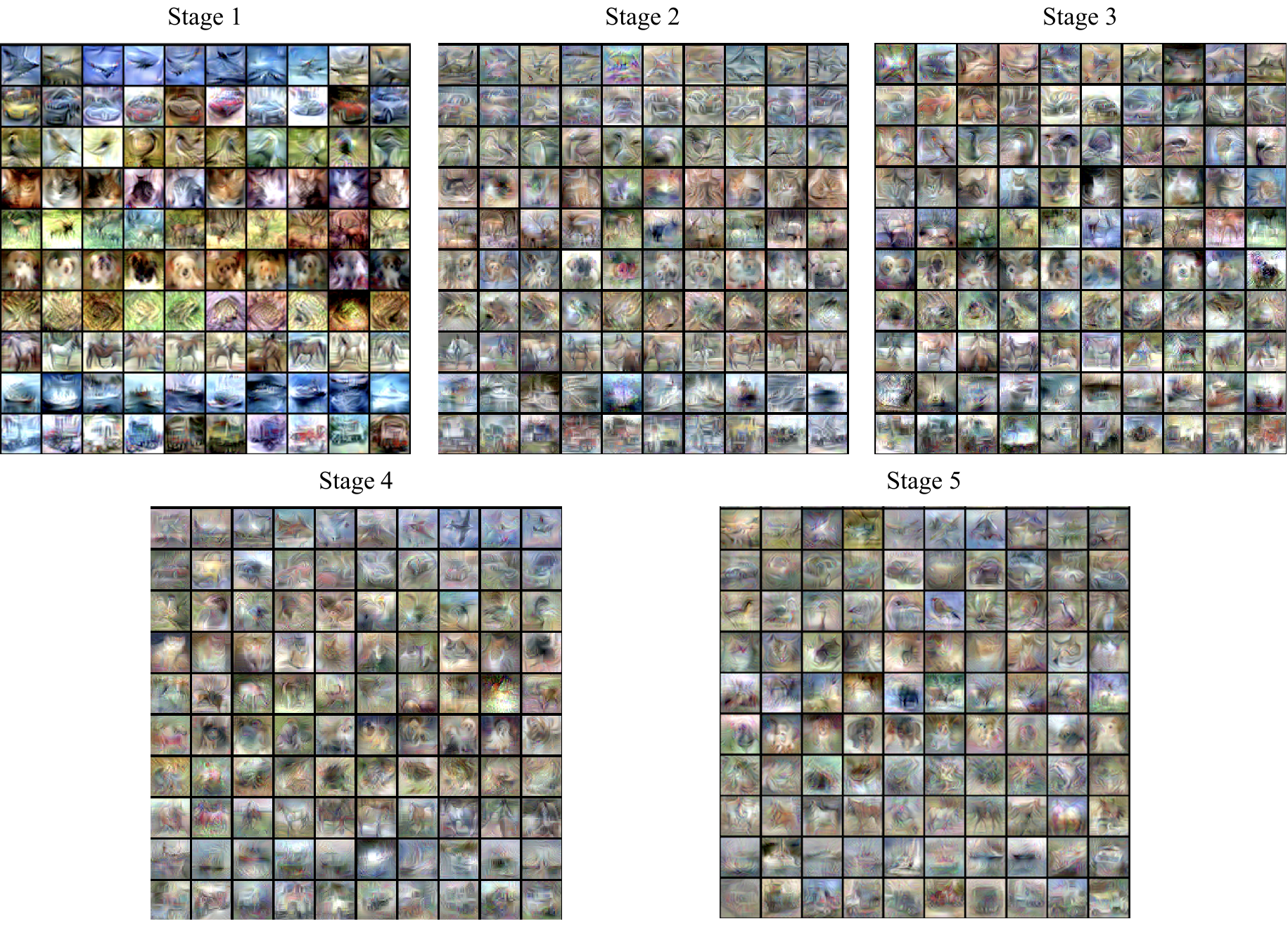}
    \caption{Visualization of synthesized samples from Stage $1$ to Stage $5$. }
    \label{fig:full_ipc10}
\end{figure}

In Figure~\ref{fig:full_ipc10} we visualize the synthetic samples of CIFAR-10 distilled at stages $1$ to $5$ using \alg+ \texttt{MTT}. We observe a significant shift of visual features in these distill images. The images distilled at the first stage are the most colorful among all the distilled samples, while the images distilled at later stages contain more abstract features and less focus on colours. These figures show that \alg helps distill diverse features according to different stages.

\section{More Experiment Results}

\subsection{Results on more methods}
We further apply \alg to \texttt{DC}~\citep{zhao2021dataset} and \texttt{DSA}~\citep{zhao2021dsa} to distill images from CIFAR-10. Table~\ref{tab:dc_dsa} shows the ConvNet's accuracy after trained on the distilled images. On \texttt{DC} and \texttt{DSA}, compared to using the single stage synthesis, \alg+ \texttt{DC} and \alg+ \texttt{DSA} generates samples that lead to higher performance, improving the baselines' performance by $2.4\%$ and $0.7\%$, respectively.

\begin{table}[ht]
    \centering
    \vspace{-1em}
    \caption{ConvNets' test accuracy on CIFAR-10 after trained on synthetic samples generated by DC and DSA with different numbers of images per class.  
    }    
    \begin{tabular}{c|c}
    \toprule  Dataset & CIFAR-10 \\
    IPC & 50\\  
   \midrule
        \texttt{DC} & $53.9\pm0.5$\\
        \texttt{DSA}  & $60.6\pm0.5$\\
        \rowcolor{lightgray!40} \texttt{PDD} + \texttt{DC} & $\textbf{56.3}\pm 0.5$ \\
        \rowcolor{lightgray!40} \texttt{PDD} + \texttt{DSA} & $\textbf{61.3}\pm0.4$ \\
        \midrule Full & {88.1} \\ \bottomrule
    \end{tabular}
    \label{tab:dc_dsa}
\end{table}

\label{sec:continual}

\end{document}